\documentclass[preprint,12pt,authoryear]{elsarticle}
\usepackage[monochrome]{xcolor}
\usepackage[linesnumbered,ruled]{algorithm2e}
\usepackage{hyperref}
 \usepackage{amsmath}
 \usepackage{float}
 \usepackage{xcolor}

 \SetKwComment{Comment}{/* }{ */}

\usepackage{amssymb}

\journal{Machine learning with applications}

\begin{document}

\begin{frontmatter}

\title{Case-Base Neural Network: survival analysis with time-varying,
higher-order interactions}

\author[inst1]{Jesse Islam}
\ead{jesse.islam@mail.mcgill.ca}
\affiliation[inst1]{organization={McGill University Department of Quantitative Life Sciences},
 addressline={805 rue Sherbrooke O}, 
 city={Montréal},
 postcode={H3A 0B9}, 
 state={Quebec},
 country={Canada}}

\author[inst2]{Maxime Turgeon}
\ead{turgeon.maxime@gmail.com}
\author[inst1,inst3]{Robert Sladek}
\ead{rob.sladek@mcgill.ca}
\author[inst4]{Sahir Bhatnagar}
\ead{sahir.bhatnagar@gmail.com}

\affiliation[inst2]{organization={University of Manitoba Department of Statistics},
 addressline={50 Sifton Rd}, 
 city={Winnipeg},
 postcode={R3T2N2}, 
 state={Manitoba},
 country={Canada}}

\affiliation[inst3]{organization={McGill University Department of Human Genetics},
 addressline={805 rue Sherbrooke O}, 
 city={Montréal},
 postcode={H3A 0B9}, 
 state={Quebec},
 country={Canada}}

\affiliation[inst4]{organization={McGill University Department of Biostatistics},
 addressline={805 rue Sherbrooke O}, 
 city={Montréal},
 postcode={H3A 0B9}, 
 state={Quebec},
 country={Canada}}
 
\begin{abstract}
\textcolor{red}{In the context of survival analysis, data-driven neural network-based methods have been developed to model complex covariate effects}. While these methods may provide better predictive performance than regression-based approaches, not all can model time-varying interactions and complex baseline hazards. To address this, we propose Case-Base Neural Networks (CBNNs) as a new approach that combines the case-base sampling framework with flexible neural network architectures. Using a novel sampling scheme and data augmentation to naturally account for censoring, we construct a feed-forward neural network \textcolor{red}{that includes} time as an input. CBNNs predict the probability of an event occurring at a given moment to estimate the \textcolor{red}{full} hazard function. We compare the performance of CBNNs to regression and neural network-based survival methods in a simulation and three case studies using two time-dependent metrics. First, we examine performance on a simulation involving a complex baseline hazard and time-varying interactions to assess all methods, with CBNN outperforming competitors. Then, we apply all methods to three real data applications, with CBNNs outperforming the competing models in two studies and showing similar performance in the third. Our results highlight the benefit of combining case-base sampling with deep learning to provide a simple and flexible framework \textcolor{red}{for data-driven modeling of single event survival outcomes that estimates time-varying effects and a complex baseline hazard by design.} An R package is available at \href{https://github.com/Jesse-Islam/cbnn}{https://github.com/Jesse-Islam/cbnn}.
\end{abstract}

\begin{highlights}
\item \textcolor{red}{Case-Base Neural Networks (CBNNs) estimate the full hazard function.}
\item \textcolor{red}{Naturally accounts for censoring and predicts smooth-in-time risk functions.}
\item \textcolor{red}{Uses a simple objective function, unlike competing methods.}
\item \textcolor{red}{Models time-varying effects by design, unlike competing methods.}
\item \textcolor{red}{CBNNs outperform the competing models in a simulation and two studies.}
\end{highlights}


\end{frontmatter}

\hypertarget{introduction}{%
\section{Introduction}\label{introduction}}

A common assumption in survival analysis is that the risk \textcolor{red}{ratio} of the event of interest does not vary with time \textcolor{red}{which has led to the dominance of proportional hazard models\citep{cox1972regression}} \citep{hanley2009}. This simplifying assumption \textcolor{red}{explains the popularity of Cox proportional hazards models over smooth-in-time, accelerated failure time (AFT) models and results in analyses} based on hazard ratios and relative risks rather than on survival curves and absolute risks \citep{hanley2009}. The proportional hazards assumption in a Cox model may be incorrect in studies where the disease pathogenesis may change over time. \textcolor{red}{For example, age, chronic obstructive pulmonary disease, diabetes mellitus, existing heart disease, prior heart surgery, intra-aortic balloon pump, urgent and emergent surgical priority are all preoperative factors that have demonstrated time-varying associations with long-term mortality following coronary artery bypass graft surgery \citep{gao2006time}. A time-varying effect of body mass index has been associated with all-cause mortality in hypertensive patients \citep{zhu2022time} and the effect of long-term statin use on type 2 diabetes onset varies with time \citep{na2020time}. The time-varying effect of tumor size has been associated with the risk of cancer recurrence \citep{coradini2000time}. Time-varying interactions were identified in appraisals of 28 of 40 targeted and immuno-oncology therapies reported by the National Institute for Health and Care Excellence, highlighting the importance of accounting for time-varying features in cancer treatment studies \citep{salmon2023clinical}.} These time-varying covariate \textcolor{red}{effects} can be incorporated easily into AFT models; however, this requires prior knowledge of potential time-varying interactions and their quantitative effects \citep{royston2002flexible}. \textcolor{red}{In this study, we develop a method that accounts for these time-varying effects without user specification, improving risk prediction models.}

We propose Case-Base Neural Networks (CBNNs) as a data-driven method for single event survival analysis. CBNNs use the case-base sampling technique, which allows probabilistic models to predict survival outcomes \textcolor{red}{by adjusting for a sampling-specific bias with an offset term} \citep{hanley2009}. After case-base sampling, we implement a model using common neural network components that use time as a feature to estimate time-varying interactions and a flexible baseline hazard.

In this paper, \textcolor{red}{we first review state-of-the-art deep learning methods for survival analysis to position CBNN relative to related works. Second,} we describe the case-base sampling \textcolor{red}{procedure} and compare its properties to neural network models, along with our hyperparameter selection \textcolor{red}{criteria}, metrics, and software implementation. \textcolor{red}{Third}, we compare the performance of CBNN to neural network and regression-based survival methods on simulated data and describe their performance in three case studies. \textcolor{red}{Finally}, we explore the implications of our results and contextualize them within neural network survival analysis in a single event setting.

 \color{red}
\section{Related works}\label{related}
 Deep learning approaches for survival analysis can be grouped into four broad categories: models that ignore censoring; Cox partial log-likelihood; discrete survival time; and fully parametric. We wish to distinguish models, where the innovation is provided by the neural network architecture, from frameworks, which alter how the survival outcome is interpreted. Based on this distinction, any framework can use any neural network architecture.

\subsection{Survival models that ignore censoring}

Survival analysis methods handle censored data, where the event time of interest may not be known. Several methods have been developed to ignore these censored individuals during the fitting process, simplifying the survival problem. For example, \citet{zadeh2020DeepSurvnet} model survival data in a deep learning context, where the model predicts survival rate categories. In contrast, two other studies predict the probability of a bad prognosis by integrating imaging and gene expression data, using either a bilinear network \citep{wang2021gpdbn} or an attention layer with transfer learning for the imaging features and ensemble forests for gene expression \citep{jia2023dccafn}. Though these methods predict a simple binary outcome, they are unable to estimate the hazard function. These architectures can be inserted into the CBNN framework which provides the added flexibility of estimating complex baseline hazards and time-varying effects. Rather than converting time-to-event outcomes into a risk-of-prognosis score, CBNN can model the full hazard function.

\subsection{Cox partial log-likelihood} \label{coxFrame}

Several architectures using the Cox partial log-likelihood have been proposed for conventional tabular datasets. \textcolor{black}{DeepSurv is a neural network-based proportional hazards model \citep{katzman2018DeepSurv}, which has been used to develop prediction models} \textcolor{red}{for several clinical scenarios \citep{ds1} \citep{ds2} \citep{ds3} \citep{she2020development}}. \citet{ching2018cox} use gene expression data as input to a single hidden layer neural network to model cancer survival; an approach which is outperformed by an autoencoder with Cox regression model \citep{yin2022cox} and a convolutional model with residual networks \citep{huang2020deep}. In parallel, \citet{hao2021deep} assess simple Cox neural networks in a high dimensional setting against random forest and penalized regression approaches. As a more complex model for gene expression, \citet{meng2022novel} propose to first fit a generative adversarial network (GAN) to the expression data before transferring the weights to a Cox neural network. These methods provide data-agnostic architectures in clinical settings that do not take advantage of known relationships between features.

 Other approaches use the prior knowledge of existing structure between features to improve predictive performance. For example,
 \citet{chen2023pathological} transfer the embeddings of pre-trained convolutional layers on histopathological images to an autoencoder, the central layer of which is passed into a linear Cox regression to predict cervical cancer survival. Histopathological images were also used to predict lung cancer survival, using either convolutional layers \citep{zhu2017wsisa} or a convolutional transformer and Siamese network model with a modified Cox partial log-likelihood loss function that accounts for aleatoric uncertainty \citep{tang2023explainable}. \citet{kaynar2023pideel} propose a pathway-informed model integrating metabolomic data and known metabolic pathways to predict survival in patients with glioma. To assess cardiovascular risk, \citet{barbieri2022predicting} propose a sex-specific neural network model using bidirectional gated recurrent units to represent recent clinical history. These architectures focus on structure within unimodal datasets and extend the concepts proposed by data-agnostic architectures.

Other approaches integrate prior knowledge across multiple modalities to improve predictive performance. \citet{hao2022joint} use sample similarity and correlations across three genomic data modalities, then integrate them in a graph neural network to model cancer outcomes. To integrate genomic and imaging data, \citet{mobadersany2018predicting} make use of convolutional layers and \citet{li2022hfbsurv} propose a factorized bilinear network as an improvement over an unfactorized network. Finally, \citet{chen2020pathomic} integrate spatial transcriptomics as well, using both convolutional layers and graphical neural networks. Whether data-agnostic, data-specific or multi-modal, all these models make use of the Cox partial log-likelihood and assume time-invariant effects.

\subsection{Discrete-time} 

Discrete-time survival models make survival predictions for specified intervals of follow-up time. For example, \textcolor{black}{DeepHit directly estimates survival probabilities and assumes an inverse Gaussian distribution as the baseline hazard \citep{lee2018DeepHit}}. The same goal is accomplished by \citet{gensheimer2019scalable} using basic hidden layers and \citet{giunchiglia2018rnn} using a recurrent neural network architecture. To estimate a piecewise hazard, \citet{kopper2022deeppamm} convert survival data into a piecewise exponential data format and propose a computationally efficient neural network. To predict cancer survival, \citet{wulczyn2020deep} incorporate prior knowledge of image motifs using convolutional layers. Discrete-time neural network models have also been used with multi-modal data. To predict long term cancer survival, \citet{vale2021long} propose a model integrating images along with transcriptomic, genomic and epigenomic data. All these methods provide a discrete estimate of survival outcomes dependent on the follow-up times of interest.

\subsection{Fully parametric framework}

 Fully parametric methods provide an estimate of the full hazard function. For example, Deep Survival Machines (DSM), uses neural networks with a mixture of distributions to fit a flexible baseline hazard \citep{dsmPaper}. This method extends to settings where the covariates are measured multiple times (i.e. time-series) \citep{nagpal2021deep}. However, the initial reports did not assess performance when time-varying effects of baseline measurements are present \citep{dsmPaper} \citep{nagpal2021deep}.

 clinical datasets often show significant time-varying effects that are relevant to disease-specific outcomes \citep{coradini2000time}\citep{gao2006time} \citep{na2020time} \citep{zhu2022time} \citep{salmon2023clinical}. Not limited to these diseases, we expect all models that predict a survival outcome may gain improved performance by accounting for non-proportional hazards. Cox partial log-likelihood based models assume the effects are time invariant \citep{cox1972regression}. CBNN accounts for the time-varying effects by design. For all the discrete-time frameworks and architectures, a selection of smaller intervals of time or more survival times of interest may result in a few or no events per interval, which potentially leads to instability in the optimization function \citep{RJ-2022-052}. In contrast, selecting larger intervals may mask nonlinear, time-varying effects in the hazard function \citep{RJ-2022-052}. CBNNs can model time-varying interactions and a complex baseline hazard function without the trade-off between long or short interval lengths. CBNNs also estimate a full hazard function for all of follow-up time, rather than discrete intervals of follow-up-time. Compared to fully parametric approaches like DSM, CBNN explicitly incorporates time into the model, providing a simple way to account for time-varying effects that improve survival prediction.


\color{black}

\hypertarget{methods}{%
\section{Case-base neural network, metrics, hyperparameters and
software}\label{methods}}

Case-base sampling is an alternate framework for survival analysis \citep{hanley2009},
which converts the total survival time into discrete \textcolor{red}{person-time coordinates} (person-moments). In this section, we detail how neural networks explicitly incorporate time as a feature while adjusting for the sampling bias in case-base sampled data\textcolor{red}{;} describe the metrics\textcolor{red}{;} and \textcolor{red}{describe the} hyperparameter selection procedure used to compare
CBNN with other methods. An R package to use CBNN is available at \url{https://github.com/Jesse-Islam/cbnn}. The entire code base to reproduce the figures and empirical results in this paper is available at \url{https://github.com/Jesse-Islam/cbnnManuscript}.

\hypertarget{case-base-sampling}{%
\subsection{Case-base sampling}\label{case-base-sampling}}

To implement case-base sampling, we divide the total survival time for each individual into discrete person-moments and treat each person-moment as a sample. This creates a \emph{base series} of \emph{person-moments} where an event does not occur. This \emph{base series} complements the \emph{case series}, which contains all person-moments at which the event of interest occurs.

For each person-moment sampled, let \(X_i\) be the corresponding covariate profile
\(\left(x_{i1},x_{i2},...,x_{ip} \right)\), \(T_i\) be the time of the person-moment and \(Y_i\) be the indicator variable for whether the event of interest occurred at time \(T_i\). We estimate the hazard function
\(h(t \mid X_i)\) using the sampled person-moments. Recall that \(h(t \mid X_i)\) is the instantaneous risk of experiencing the event at time \(t\) for a given set of covariates \(X_i\), assuming \(T_i \geq t\).

Now, let \(b\) be the (user-defined) size of the \emph{base series} and let \(B\) be the sum of all follow-up times for the individuals in the study. Let $c$ be the number of events in the \emph{case series}. A reasonable concern is determining how large $b$ should be relative to $c$, since the size of \(b\) influences how much information is lost in the sampling process \citep{hanley2009}. The relative information when comparing two averages is measured by $\frac{cb}{c+b}=\left[\left( \frac{1}{c}+\frac{1}{b}\right)\right]$, where $b$ is the sample size of the \emph{base series} and $c$ is the sample size of the \emph{case series} \citep{hanley2009} \citep{mantel1}. If $b=100c$, then we expect learned weights to be at most one percent higher than if the entire study base $B$ was used, as they are proportional to $\frac{1}{c}+\frac{1}{100c}$ rather than $\frac{1}{c}+\frac{1}{\infty c}$ \citep{hanley2009} \citep{mantel1}.

If we sample the \emph{base series} uniformly across the study base, then the hazard function of the sampling process is equal to \(b/B\). Therefore, we have the following equality \citep{saarela2015}
[For a rigorous treatment, see Saarela \& Hanley (2015) section 3]:
\begin{align}\label{eqn:main}
\frac{P\left(Y_i=1 \mid X_i, T_i\right)}{P\left(Y_i = 0 \mid X_i, T_i\right)} = \frac{h\left(T_i \mid X_i\right)}{b/B}.
\end{align} The odds of a person-moment being in the \emph{case
series} is the ratio of the hazard \(h(T_i \mid X_i)\) and the uniform
rate \(b/B\). Using \eqref{eqn:main}, we can see how the log-hazard
function can be estimated from the log-odds arising from case-base
sampling: \begin{align}\label{eqn:offset}
\log \left( h\left(t \mid X_i\right)\right) = \log \left(\frac{P\left(Y_i = 1 \mid X_i, t\right)}{P\left(Y_i = 0 \mid X_i, t\right)}\right) + \log\left(\frac{b}{B}\right).
\end{align}

To estimate the correct hazard function, we adjust for the bias introduced when sampling a fraction of the study base (\(\frac{b}{B}\)) by adding the term \(\log\left(\frac{B}{b} \right)\) to offset \(\log\left(\frac{b}{B} \right)\) during the fitting process. \textcolor{red}{See Algorithm \ref{alg:1} for a pseudocode implementation of the case-base sampling procedure. For a discussion of the connection between our objective function and the full data likelihood, see \cite{saarela2016case}.}

\begin{algorithm}
\color{red}
 \SetKwInOut{Input}{Input}
 \SetKwInOut{Output}{Output}
 \underline{function sampleCaseBase$(X)$}\;
 \Input{X = matrix of covariate profiles. \\ $X_{T}$ = Vector of event times for each profile. \\ $X_{Y}$ = Vector of event indicators for each profile.}
 \Output{Case-base sampled data matrix.}
 $n$ = length($X_{T}$) \Comment*[r]{Number of subjects}
 $B$ = Sum($X_{T}$) \Comment*[r]{Total person-time in study base}
 $c$ = Sum($X_{Y}$) \Comment*[r]{Number of cases}
 $b$ = 100 * c \Comment*[r]{Number of base series samples}
 \For{$i = 1, \ldots, n$}{
 Set $p_i = X_T[i]/B$\;
}
\While{$j < b$}{
 Sample $\ell$ from $\{1, \ldots, n\}$ with probabilities $p_1, \ldots, p_n$\\
 Set $baseSeries[j, ] = X[\ell, ]$\\
 Sample $baseSeries_T[j]$ from $\mathrm{Uniform}(0, X_t[\ell])$\\
}
 Set $caseSeries = X[X_{Y}==1, ]$.\\
 return(concatenate($baseSeries$,$caseSeries$))
 \caption{Pseudocode algorithm of the case-base sampling procedure.}\label{alg:1}
\end{algorithm}

\color{black}

\hypertarget{neural-networks-to-model-the-hazard-function}{%
\subsection{Neural networks to model the hazard
function}\label{neural-networks-to-model-the-hazard-function}}

\begin{figure}

{\centering \includegraphics[width=1\linewidth]{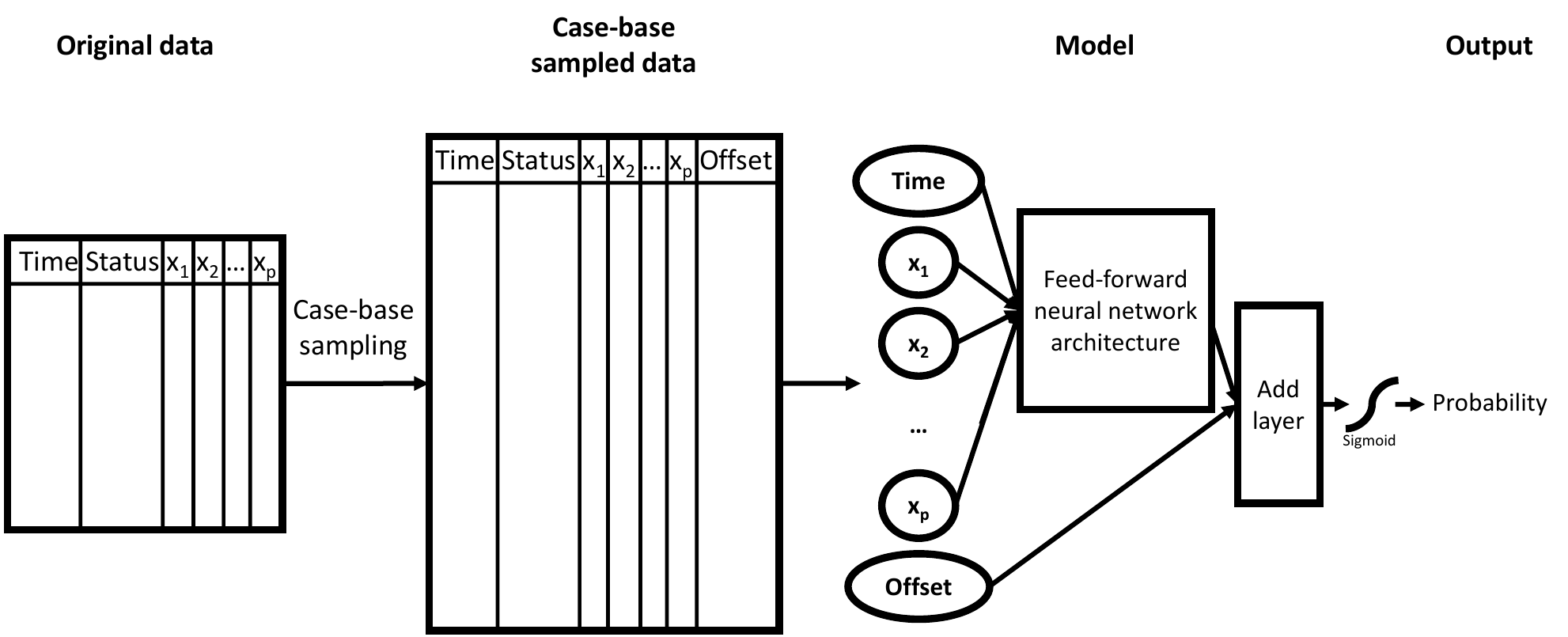}}

\caption{Methodological steps involved in CBNN. The first step, case-base sampling, is completed before training begins. Then, we pass this sampled data through a feed-forward neural network\textcolor{red}{,} add an offset to adjust for the bias inherent in case-base sampling and apply a sigmoid activation function to \textcolor{red}{estimate} a probability. Once the neural network model completes its training, we can convert the probability to a hazard for the survival outcome of interest.}\label{fig:NNarch}
\end{figure}

After case-base sampling, we pass all features, including time, into any user-defined feed-forward component (Figure \ref{fig:NNarch}). Then, we add an offset term and pass the output through a sigmoid activation function (Figure \ref{fig:NNarch}). Since we are interested in predicting the odds of an event occurring, the sigmoid activation function is ideal as it is the inverse of the odds and can be used to calculate the hazard. The general form for \textcolor{red}{any feed-forward} neural network \textcolor{red} {architecture} using CBNN is:

\begin{align*}
P\left(Y=1|X,T\right)&=\mathrm{sigmoid}\left(f_{\theta}(X, T) + \log\left(\frac{B}{b}\right) \right),
\end{align*}

\noindent where \(T\) is a random variable representing the event time, \(X\) is the random variable for a covariate profile, \(f_{\theta}(X, T)\) represents any feed-forward neural network architecture, \(\log\left(\frac{B}{b}\right)\) is the offset term to adjust for the bias (\(\log\left(\frac{b}{B}\right)\)) set by case-base sampling, \(\theta\) is the set of parameters learned by the neural network and \(\mathrm{sigmoid}(x)=\frac{1}{1+e^{-x}}\). By approximating a higher-order polynomial of time using a neural network, the baseline hazard specification is now data-driven, while user-defined hyperparameters such as regularization, number of layers and nodes control the flexibility of the hazard function.

The following derivation shows how our probability estimate is converted
to odds: \begin{align*}
 \log\left( h(t \mid X) \right) &= \log\left(\frac{\mathrm{sigmoid}\left(f_{\theta}(X, T) + \log\left(\frac{B}{b}\right)\right)}{1-\mathrm{sigmoid}\left(f_{\theta}(X, T) + \log\left(\frac{B}{b}\right)\right)}\right) + \log\left(\frac{b}{B}\right) \\
 &= \log\left( \frac{\frac{\exp\left(f_{\theta}(X, T) + \log\left(\frac{B}{b}\right)\right)}{\exp\left(f_{\theta}(X, T) + \log\left(\frac{B}{b}\right)\right)+1}}{1-\frac{\exp\left(f_{\theta}(X, T) + \log\left(\frac{B}{b}\right)\right)}{\exp\left(f_{\theta}(X, T) + \log\left(\frac{B}{b}\right)\right)+1}}\right) + \log\left(\frac{b}{B}\right) \\
 &= \log\left(\exp\left( f_{\theta}(X, T) + \log\left(\frac{B}{b}\right) \right) \right) + \log\left(\frac{b}{B}\right) \\
 &= f_{\theta}(X, T) + \log\left(\frac{B}{b}\right) + \log\left(\frac{b}{B}\right) \\
&= f_{\theta}(X, T). 
\end{align*}

We use binary cross-entropy as our loss function \citep{gulli2017}\textcolor{red}{, from which we calculate}:
\begin{align*}
L(\theta)=-\frac{1}{N}\left( \sum^{N}_{i=1} y_{i} \cdot \log(\hat{f}_{\theta}(x_{i}, t_{i}) ) + (1-y_{i} )\cdot \log(1-\hat{f}_{\theta}(x_{i}, t_{i}) ) \right),
\end{align*} where \(\hat{f}_{\theta}(x_{i}, t_{i})\) is our estimate for a given covariate profile and time, \(y_{i}\) is our target value specifying whether an event occurred, and \(N\) represents the number of individuals in our training set.

Backpropagation with an appropriate minimization algorithm (such as Adam, RMSPropagation, stochastic gradient descent) is used to optimize the parameters in the model \citep{gulli2017}. For our analysis, we use Adam \textcolor{red}{\citep{kingma2014adam}}. While the size of the \emph{case series} is fixed as the number of events; the size of the \emph{base series} is not restricted. We use a ratio of 100:1 \emph{base series} to \emph{case series} \citep{hanley2009}. After fitting our model, we convert the output to a hazard. To use CBNN for predictions, we manually set the offset term $\left(\log\left(\frac{B}{b} \right)\right)$ to 0 in the new data as we already account for the sampling bias during the fitting process.

Since we are directly modeling the hazard, we can readily estimate the risk function (\(F\)) at time \(t\) for a covariate profile \(X\),
\begin{align}\label{eqn:ci2}
F\left(t\mid X\right)& = 1 - \exp\left(-\int_{0}^{t}h(u|X) \,\textrm du\right).
\end{align} We use a finite Riemann sum \citep{hughes2020calculus} to approximate the integral in \eqref{eqn:ci2}.

\hypertarget{performance-metrics}{%
\subsection{Performance metrics}\label{performance-metrics}}

To choose our method-specific hyperparameters, we use the Integrated Brier Score (IBS) \citep{graf1999}, which is based on the Brier Score (BS) and provides a summarized assessment of performance for each model. We assess the performance of models on a held-out dataset using two metrics: 1) the Index of Prediction Accuracy (IPA) \citep{kattan2018index}; and 2) the Inverse probability censoring weights-adjusted time-dependent area under the receiver operating characteristic curve ($AUC_{IPCW}$) \citep{auc}. The IPA score is a metric for both discrimination and calibration, while the $AUC_{IPCW}$ provides a metric for discrimination only.

\hypertarget{bs}{%
\subsubsection{Brier Score (BS)}\label{bs}}
The BS \citep{graf1999} is defined as \begin{align}\label{eqn:bs}
\resizebox{0.9\hsize}{!}{$%
BS(t)=\frac{1}{N}\sum^{N}_{i=1}\left(\frac{\left(1 - \widehat{F}(t \mid X_{i})\right)^{2}\cdot I(T_{i}\leq t,\delta_{i}=1)}{\widehat{G}(T_{i})} + \frac{\widehat{F}(t\mid X_{i})^{2}\cdot I(T_{i}>t)}{\widehat{G}(t)}\right),
$%
}%
\end{align} where \(\delta_{i}=1\) shows individuals who have experienced the event, \(N\) represents the number of samples in our dataset over which we calculate \(BS(t)\), \(T_{i}\) is the survival or censoring time of an individual, \(\widehat{G}(t)=P[c>t]\) is a non-parametric estimate of the censoring distribution and \(c\) is censoring time. The BS provides a score that accounts for the information loss because of censoring. Once we fix our \(t\) of interest, the individuals in the dataset can be divided into three groups. Individuals who experienced the event before \(t\) are present in the first term of the equation. The second term of the equation includes individuals who experience the event or are censored after \(t\). Those censored before \(t\) (the remaining individuals) are accounted for by the IPCW adjustment (\(G(\cdot)\)) present on both terms.

\hypertarget{integrated-brier-score-ibs}{%
\subsubsection{Integrated Brier Score (IBS)}\label{integrated-brier-score-ibs}}

The Integrated Brier Score (IBS) is a function of the BS (\ref{eqn:bs}) \citep{graf1999}, which is defined as \begin{align*}
IBS(t)=\int_{t}^{t_{max}}BS(t)dw(t),
\end{align*} where $w(t)=\frac{t}{t_{max}}$ and $t_{max}$ is the upper bound for the survival times of interest. As the IBS is calculated for a range of follow-up times, it is a useful metric to track overall performance across all of follow-up time during cross-validation. We use the IPA score to get an estimate of performance at each follow-up time for our studies.

\hypertarget{index-of-prediction-accuracy-ipa}{%
\subsubsection{Index of prediction accuracy (IPA)}\label{index-of-prediction-accuracy-ipa}}

The IPA is a function of time, based on the BS. The IPA score is given by \begin{align}
\textrm{IPA}(t) &= 1-\frac{BS_{model}(t)}{BS_{null}(t)}, \nonumber
\end{align} where $BS_{model}(t)$ represents the BS over time $(t)$ for the model of interest and $(S_{null}(t)$ represents the BS if we use an unadjusted Kaplan-Meier (KM) curve as the prediction for all observations \citep{kattan2018index}. The IPA score has an upper bound of one, where positive values show an increase in performance over the null model and negative values show that the null model performs better. As an extension of BS, the IPA score assesses both model calibration and discrimination and shows how performance changes over follow-up time \citep{graf1999} \citep{kattan2018index}.

\hypertarget{auc2}{%
\subsubsection{Inverse probability censoring weights-adjusted time-dependent area under the receiver operating characteristic curve (\texorpdfstring{$AUC_{IPCW}$}{AUC\_IPCW})}\label{auc2}}

The IPCW-adjusted AUC ($AUC_{IPCW}$) is a time-dependent metric that considers censoring \citep{auc}. For a given follow-up time of interest,

\begin{equation*}
\resizebox{\hsize}{!}{$AUC_{IPCW}(t)=\frac{\sum^{n}_{i=1}\sum^{n}_{j=1} I_{\widehat{F}(t|X_{i})>\widehat{F}(t|X_{j})} \cdot I_{T_{i}\leq t,\delta_{i}=1} \cdot \left( 1-I_{T_{j}\leq t,\delta_{j}=1} \right) \cdot W_{i}(t) \cdot W_{j}(t)} { \sum^{n}_{i=1}\sum^{n}_{j=1} I_{T_{i}\leq t,\delta_{i}=1} \cdot \left( 1-I_{T_{j}\leq t,\delta_{j}=1} \right) \cdot W_{i}(t) \cdot W_{j}(t) },$}
\end{equation*}

where \begin{align*}
W_{i}(t)=\frac{I_{T_{i}\leq t,\delta_{i}=1}}{\widehat{G}(T_{i})} +\frac{I_{T_{i}>t}}{\widehat{G}(t)}.
\end{align*} The $AUC_{IPCW}$ measures discrimination \citep{auc}. By examining both IPA and $AUC_{IPCW}$,
we can better understand whether a model performs better in terms of calibration or discrimination.

\hypertarget{hyperparameter-selection}{%
\subsection{Hyperparameter selection}\label{hyperparameter-selection}}

Neural networks require external parameters that constrain the learning process (hyperparameters). We apply the following hyperparameter optimization procedure to each method for each study. First, we fix a test set with 15\% of the data which we keep aside during hyperparameter selection. To determine the best hyperparameters for each neural network method, we use a three-fold cross-validated grid search on the remaining data (85\% training, 15\% validation) for the following range of hyperparameters:
\[
\textrm{Search space}: \begin{cases}
\textrm{Learning rate} \sim \{0.001, 0.01\} \\ 
\textrm{Dropout} \sim \{0.01,0.05, 0.1\} \\
\textrm{First layer nodes} \sim \{50, 75, 100\} \\
\textrm{Second layer nodes} \sim \{10,25,50\} \\
\textrm{Number of batches} \sim \{100, 500\} \\
\textrm{Activation function} \sim \{\textrm{ReLU, Linear}\}\\
\alpha \sim \{0, 0.5, 1\} \ (\textrm{DeepHit only}).
\end{cases}
\]

\textcolor{red}{For all methods, we use the training set to estimate the mean and standard deviation and scale each covariate of interest other than time, which is divided by the training set maximum. For CBNN, we perform case-base sampling independently for the training and validation sets, then use the estimated training offset as our validation offset. The linear models combine the training set and validation set for training.} DeepHit uses $\alpha$ as a hyperparameter, while DeepSurv and CBNN do not. We track IBS on the validation set for each hyperparameter combination, choosing the combination with the lowest score for each method (Table \ref{tab:wins}).

\begin{table}[H]
\caption{Hyperparameters selected after three-fold cross-validated grid search along with the average IBS for each neural network model in the complex simulation (A), multiple myeloma (MM) case study (B), free light chain (FLC) case study (C) and prostate cancer (Prostate) case study (D).}
\label{tab:wins}
\begin{center}\includegraphics[width=1\linewidth]{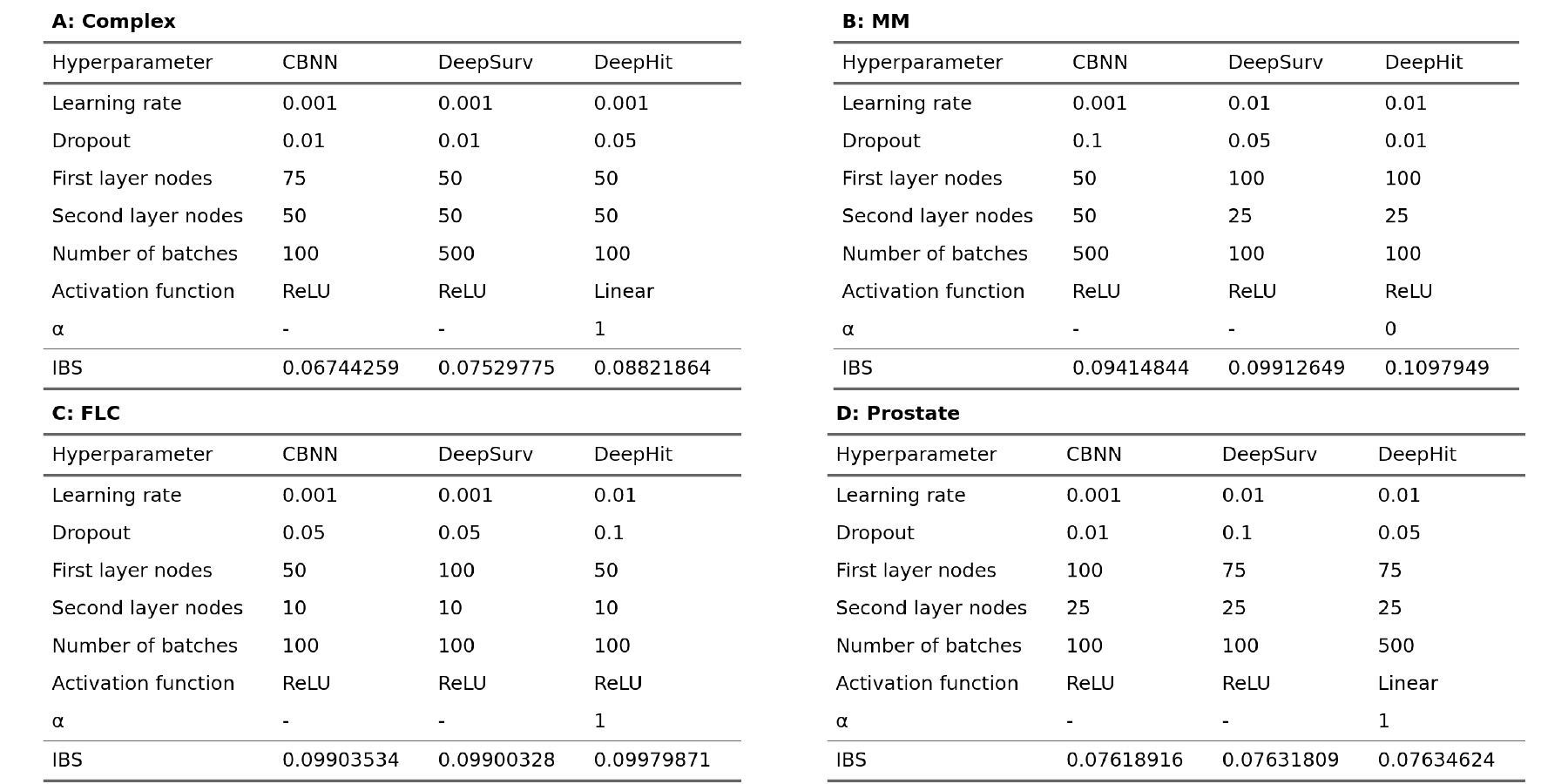} \end{center}
\end{table}

\hypertarget{software-implementation}{%
\subsection{Software implementation}\label{software-implementation}}

R \citep{Rsoft} and Python \citep{py} are used to evaluate methods from both languages. We fit Cox model\textcolor{red}{s} using the \textbf{survival} package
\citep{survpkg} and CBLR model\textcolor{red}{s} using the \textbf{casebase} package \citep{cbpkg}. Both DeepSurv and DeepHit are fit using \textbf{pyCox}
\citep{lee2018DeepHit}. We made the components of CBNN using the \textbf{casebase} package \citep{cbpkg} for the sampling step and the
\textbf{keras} package \citep{keras} for our neural network architecture. We use the \textbf{simsurv} package \citep{simsurv} for our simulation studies
and \textbf{flexsurv} \citep{flexsurv} to fit a flexible baseline hazard using splines for our complex simulation. The \textbf{riskRegression} package
\citep{riskRegression} is used to get the IPA and $AUC_{IPCW}$. We modify the \textbf{riskRegression} package to be used with any user supplied
risk function \(F\). We use the \textbf{reticulate} package \citep{reticulate} to run both R and Python based methods on the same seed.


\begin{figure}[H]

{\centering \includegraphics[width=1\linewidth]{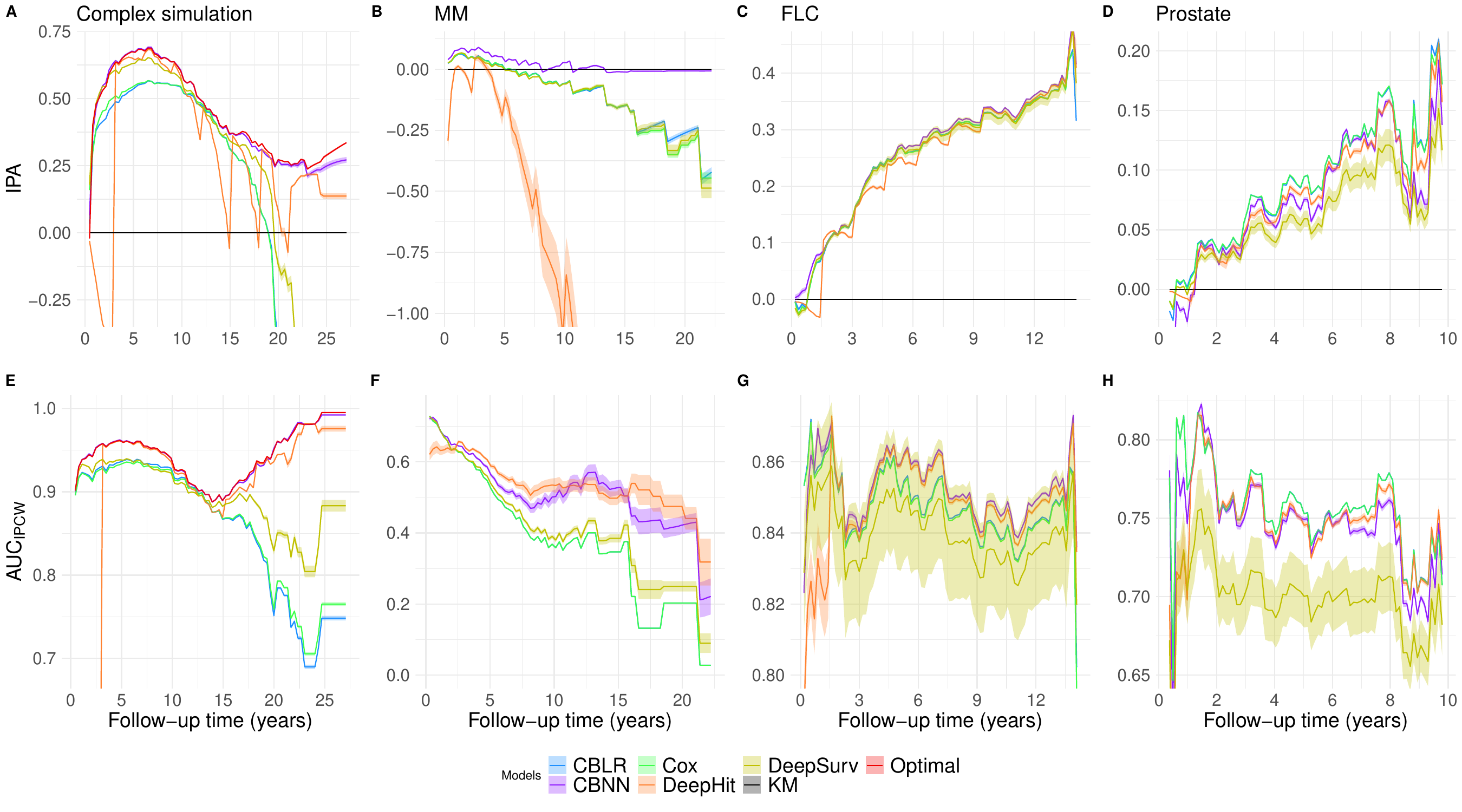} 

}

\caption{Performance of each model in the complex simulation (A, E), multiple myeloma (MM) case study (B, F), free light chain (FLC) case study (C, G) and prostate cancer (Prostate) case study (D, H). The first row shows the IPA for each model in each study over follow-up time. Negative values mean the model performs worse than the null model and positive values mean the model performs better. The second row shows the $AUC_{IPCW}$ for each model in each study over follow-up time, where higher is better. Each model-specific metric in each study shows a 95\% confidence interval over 100 iterations. Metrics are shown for six models: Case-Base with Logistic Regression (CBLR), Case-Base Neural Network (CBNN), Cox Proportional Hazard (Cox), DeepHit and DeepSurv. The Kaplan-Meier (KM) model serves as a baseline, predicting the average curve for all individuals. CBLR and Cox have near identical performance, resulting in curves that overlap. The Optimal model (a CBLR model with the exact interaction terms and baseline hazard specified) shows the best performance we can expect on the simulated data.}\label{fig:megaPlot}
\end{figure}

\hypertarget{sims}{%
\section{Simulation study}\label{sims}}

We simulate data to evaluate the performance of CBNN in comparison with existing regression (Cox, CBLR) and neural network (DeepHit, DeepSurv) methods. We specify a linear combination of each covariate as the linear predictor in the regression-based methods (Cox, CBLR), which contrasts with neural network approaches that allow for approximations of non-linear interactions. We simulate data based on a complex baseline hazard with time-varying interactions and 10\% random censoring. We simulate three covariates for 5000 individuals:

\begin{equation*}
\resizebox{\hsize}{!}{$
z_{1} \sim \textrm{Bernoulli}(0.5) \qquad \qquad 
z_{2} \sim \begin{cases}
 N(0,0.5) & \textrm{if } z_{1}=0\\ 
 N(1,0.5) & \textrm{if } z_{1}=1
\end{cases} \qquad \qquad
z_{3} \sim N(1,0.5).$}
\end{equation*}

In addition to the methods described above, we include the exact functional form of the covariates in a CBLR model (referred to as Optimal for simplicity) in the complex simulation. We obtain confidence intervals by conducting 100 bootstrap re-samples on the training data. We keep 15\% of the data for testing before hyperparameter selection. We use 15\% of the remaining data for validation and the rest is reserved for training. \textcolor{red}{We do not perform case-base sampling on the testing set and we set the offset term to 0 during prediction because the bias from case-base sampling is appropriately adjusted for during the fitting process.} We predict risk functions for individuals in the test set, which are used to calculate our IPA and $AUC_{IPCW}$.

\hypertarget{complex-simulation-flexible-baseline-hazard-time-varying-interactions}{%
\subsection{Complex simulation: flexible baseline hazard, time-varying interactions}\label{complex-simulation-flexible-baseline-hazard-time-varying-interactions}}

We use this simulation to assess performance on data with a complex baseline hazard and a time-varying interaction. We design the model \begin{align}
\log h(t \mid X_i) =\sum_{i=1}^{5} (\gamma_{i} \cdot \psi_{i}) + \beta_{{1}} (z_{1}) + \beta_{{2}} (z_{2})+ \beta_{{3}} (z_{3})+ \tau_{1} ( z_{1} \cdot t)+ \tau_{2} (z_{2} \cdot z_{3}), \nonumber
\end{align} where
\(\gamma_{1}=3.9, \gamma_{2}=3, \gamma_{3}=-0.43, \gamma_{4}=1.33,\gamma_{5}=-0.86, \beta_{{1}}=-5, \beta_{{2}}=-1, \beta_{{3}}=1, \tau_{1}=0.001, \tau_{2}=-1\) and \(\psi_{i}\) are basis splines. The \(\gamma\) coefficients are obtained from an intercept-only cubic splines model with three knots using the \emph{flexsurvspline} function from the \textbf{flexsurv} package \citep{flexsurv} on the German Breast Cancer Study Group dataset. \textcolor{red}{These coefficients provide parameters for} a complex baseline hazard from which we can simulate. The study comprised \textcolor{red}{of} 686 women with breast cancer followed between 1984 and 1989 \citep{royston2002flexible}. These $\gamma$, $\beta$ and $\tau$ coefficients are used as our baseline hazard parameters and are fixed for the analysis. The \(\beta\) coefficients represent direct effects, \(\tau_{2}\) represents an interaction and \(\tau_{1}\) is a time-varying interaction.

\hypertarget{performance-comparison-in-complex-simulation}{%
\subsection{Performance comparison in complex
simulation}\label{performance-comparison-in-complex-simulation}}

Figure \ref{fig:megaPlot} A, E and Table \ref{tab:megaTable} shows the performance over time on a test set. The Optimal model acts as a reference for ideal performance on the simulated data. For discrimination, the Optimal model performs best, followed by CBNN, DeepHit, DeepSurv and the linear models (Figure \ref{fig:megaPlot} E). To obtain a more realistic performance assessment, we compared models in three case studies with a time-to-event outcome.

\begin{table}[!htbp]
\caption{Performance at percentages of follow-up time in the complex simulation (A), multiple myeloma (MM) case study (B), free light chain (FLC) case study (C) and prostate cancer (Prostate) case study (D). Each table shows performance for each method at $25\%$, $50\%$, $75\%$ and $100\%$ of follow-up time. The models of interest are case-base with logistic regression (CBLR), Cox, Case-Base Neural Network (CBNN), DeepHit, DeepSurv, and Optimal (in the complex simulation). The best score at each percent of follow-up time is highlighted in bold. If tied, then all tied values are highlighted.}
\label{tab:megaTable}
\begin{center}\includegraphics[scale=0.225]{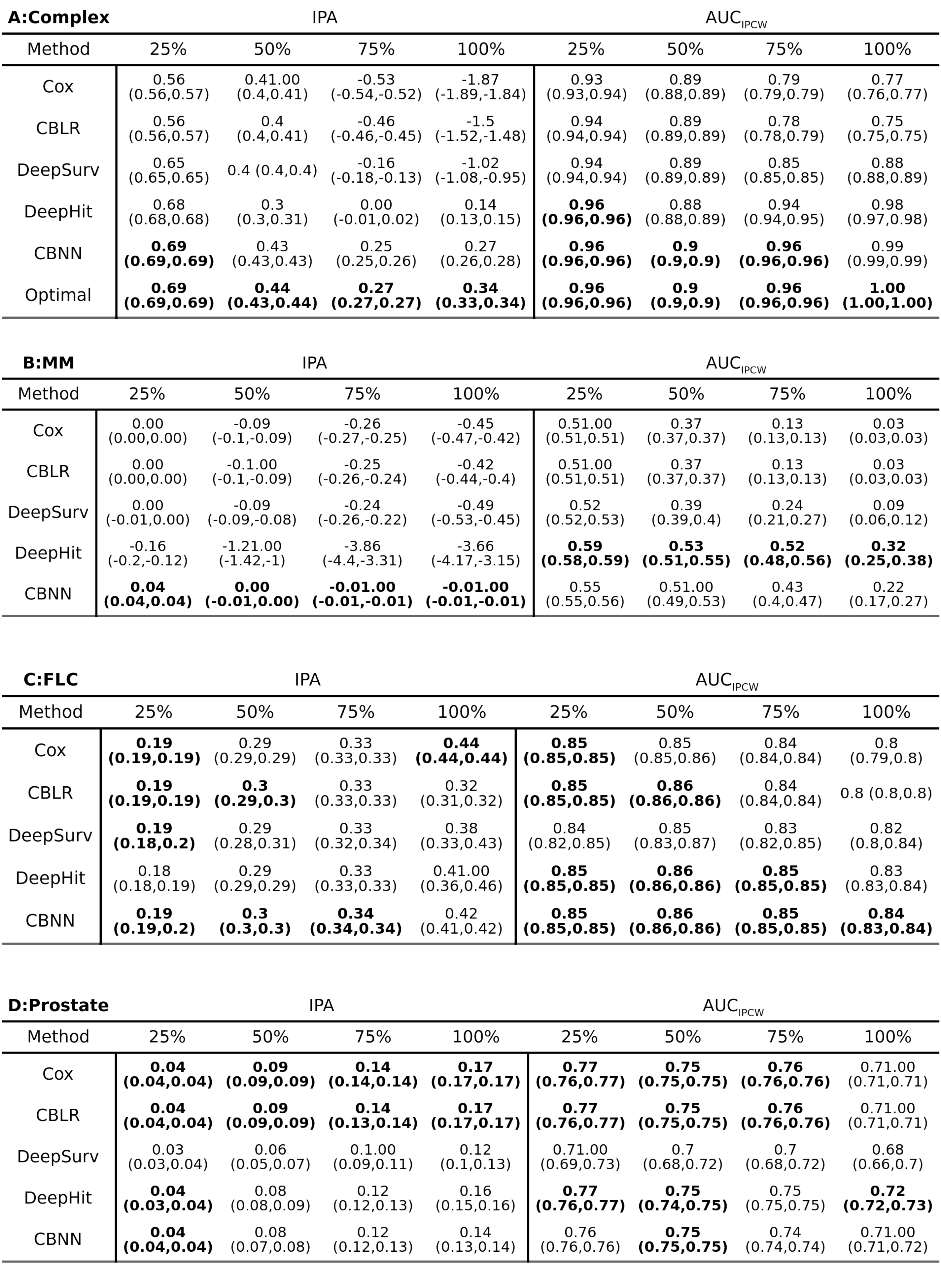} \end{center}
\end{table}


\hypertarget{casestudies}{%
\section{Case studies}\label{casestudies}}

The simulation examines whether a complex baseline hazard and time-varying interactions affect method performance. The case study assesses performance in more realistic conditions, where we may not know if a flexible baseline hazard or time-varying \textcolor{red}{interactions are beneficial for prediction}. For each dataset, we perform a grid search with the same hyperparameter procedure as described in the simulation (three-fold cross-validated grid search). We then predict risk functions for everyone in the test set with the selected hyperparameters, which is used to calculate our metrics. We conduct 100-fold bootstrap re-samples on the training data to get confidence intervals.

\hypertarget{pe-multiplemyeloma}{%
\subsection{Performance evaluation on multiple myeloma dataset}\label{pe-multiplemyeloma}}
We expect cancer may have risk factors that vary with time \citep{coradini2000time} \textcolor{red}{\citep{salmon2023clinical}}. As such, the first case study examines multiple myeloma (MM), using a cohort of 3882 patients seen at the Mayo Clinic from 1947 to 1996 until death \citep{myeloma}. Data are provided by the \textbf{survival} package \citep{survpkg}. We use two covariates, year of entry into the study and the time of MM diagnosis \citep{myeloma}. We see 71\% incidence over 23 years \citep{myeloma}.

Figure \ref{fig:megaPlot} B, F and Table \ref{tab:megaTable} B demonstrate the performance over time on a test set. For discrimination, CBNN and DeepHit have a similar performance, followed by DeepSurv and finally the linear models performing worst (Figure \ref{fig:megaPlot} F). For both discrimination and calibration, CBNN performs substantially better than the linear models and DeepSurv, with DeepHit having the worst performance overall (Figure \ref{fig:megaPlot} B). Together, CBNN is the best calibrated model and one of the best at discrimination in the MM case study.


\hypertarget{pe-flc}{%
\subsection{Performance evaluation on free light chain dataset}\label{pe-flc}}
Serum free light chain (FLC) is a known diagnostic tool for assessing MM \citep{mm2flc}. We are interested in whether there is a predictive benefit if the FLC markers varies with time. As such the second case study examines the relationship between serum FLC and mortality in a random sample of half the individuals in $\frac{2}{3}$ of the residents of Olmsted County over the age of 50 \citep{flc}. Data are provided by the \textbf{survival} package \citep{survpkg} for 7874 subjects tracked until death with 27\% incidence over 14 years \citep{flc}. We use five covariates, total serum FLC (sum of kappa and lambda), age, sex, serum creatine and monoclonal gammopathy state \citep{flc}.

Figure \ref{fig:megaPlot} C, G and Table \ref{tab:megaTable} C demonstrate the performance over time on a test set. CBNN, DeepHit and the linear models perform best at discrimination, followed by DeepSurv (Figure \ref{fig:megaPlot} G). The rankings remain the same aside from DeepHit where the performance periodically drops to worse than DeepSurv (Figure \ref{fig:megaPlot} C). Together, CBNN outperforms the competing models in terms of both calibration and discrimination. However, CBNN is only slightly better than the linear models.

\hypertarget{pe-prostate}{%
\subsection{Performance evaluation on prostate cancer dataset}\label{pe-prostate}}
As the CBNN model saw large predictive benefits on MM and small predictive benefits on FLC, we wish to examine longitudinal data with \textcolor{red}{another} complex risk profile.
The third case study (Prostate) examines prostate cancer survival on a publicly available simulation of the Surveillance, Epidemiology, and End Results (SEER) medicare study \citep{prostate}.
The Prostate dataset, which is contained in the \textbf{asaur} package \citep{asaur}, has a record of competing risks. As we are only interested in the single event scenario, we only keep individuals with prostate cancer death or censoring. This subset tracks 11054 individuals with three covariates: differentiation grade, age group and cancer state. There is a 7\% incidence over 10 years \citep{prostate}.

Figure \ref{fig:megaPlot} C, G and Table \ref{tab:megaTable} C demonstrate the performance over time on a test set. In the prostate case study, the linear models outperform the other models in both discrimination and calibration, followed by CBNN and DeepHit and finally DeepSurv (Figure \ref{fig:megaPlot} C, G). Aside from DeepSurv, the performance is similar across all models.


\hypertarget{discussion}{%
\section{Discussion}\label{discussion}}

CBNNs model survival outcomes by using neural networks on case-base sampled data. We incorporate follow-up time as a feature, providing a data-driven estimate of a flexible baseline hazard and time-varying interactions in our hazard function. The two competing neural network models we evaluated cannot model time-varying interactions \citep{katzman2018DeepSurv} \citep{lee2018DeepHit}. The CBNN model performs better due to its ability to model time-varying interactions and a complex baseline hazard.

The complex simulation requires a method that can learn both time-varying interactions and have a flexible baseline hazard. Based on our complex simulation results (Figure \ref{fig:megaPlot} A, E and Table \ref{tab:megaTable} A), CBNN outperforms the competitors. This simulation shows how all models perform under ideal conditions with minimal noise in the data, while the three case studies assess their performance in realistic conditions. In the MM case study, flexibility in both interaction modeling and baseline hazard improves the performance of CBNN over the other models, suggesting that this flexibility aids calibration (Figure \ref{fig:megaPlot} B, F and Table \ref{tab:megaTable} B). Upon examination of the FLC case study, CBNN demonstrates a small improvement to performance compared to the linear models and DeepHit for both IPA and AUC (Figure \ref{fig:megaPlot} C, G and Table \ref{tab:megaTable} C). In the Prostate case study, the linear models outperform the neural network ones, while CBNN and DeepHit alternate their positions depending on the follow-up time of interest and DeepSurv maintains last place (Figure \ref{fig:megaPlot} C, G and Table \ref{tab:megaTable} C). We attribute this to potential over-parameterization in the neural network models, as we did not test for fewer nodes in each hidden layer, even with dropout. Though the ranking places the linear models above the neural network ones, their overall performance falls within a small range of IPA and AUC values aside from DeepSurv.

Compared to CBNN, the neural network competitors have limitations. DeepSurv is a proportional hazards model and does not estimate the baseline hazard \citep{katzman2018DeepSurv}. DeepHit requires an alpha hyperparameter, assumes a single distribution for the baseline hazard and models the survival function directly \citep{lee2018DeepHit}. The alternative neural network methods match on time, while CBNN models time directly. \textcolor{red}{By modeling time directly, CBNN can approximate a flexible baseline hazard with whichever time-varying effects may be relevant to the outcome of interest.}

While we apply a full grid search with three-fold cross-validation for all neural network models, there are an infinite number of untested hyperparameters we did not test. A completely exhaustive search is computationally infeasible; therefore, it is reasonable to expect that there exists a set of hyperparameters that is better suited for each model in each study. However, we test a reasonably large range while accounting for potential over-fitting by including dropout \citep{srivastava2014dropout}. We provide the same options to all models aside from DeepHit, which has a method specific hyperparameter $\alpha$ \citep{lee2018DeepHit}. \textcolor{red}{Another method of particular interest is DSM, as it is a fully parametric model \citep{dsmPaper}. However,} with our hyperparameter options, \textcolor{red}{DSM} did not converge in the complex simulations, which may have been due to issues in software or method specific limitations. Therefore, we did not include this method in our comparison. With these limitations in mind, we summarize the differences in performance across all methods in our comparisons. 

If prediction is our goal, we suggest CBNN as the best model in the Complex simulation, MM case study and FLC case study. The linear models were competitive in the FLC case study and performed best in the Prostate case study. Though neural network interpretability is steadily improving \citep{interpret}, there is still a trade-off compared to regression models, especially when the predictive performance gain is minimal, like in the FLC case study. We suggest that reference models should be included when assessing neural network models. Both a null model (KM curve) and a linear model (either Cox or a flexible baseline hazard model like CBLR) provide insight as to whether the neural network model is learning anything useful beyond linear predictors that were not accounted for.

\section{Conclusions}\label{sec5}

Our study was motivated by the lack of easily implemented survival models based on neural networks that can approximate time-varying interactions. This led us to apply the case-base sampling technique, which has previously been used with logistic regression \citep{hanley2009}, to neural network models for a data-driven approach to time-varying interaction modeling. In our paper, \textcolor{red}{we} aim to compare CBNNs with existing methods. We assess the discrimination and calibration of each method in a simulation with time-varying interactions and a complex baseline hazard, and three case studies. CBNNs outperform all competitors in the complex simulation and two case studies while maintaining competitive performance in a final case study. Once we perform case-base sampling and adjust for the sampling bias, we can use a sigmoid activation function to predict our hazard function. Our approach provides an alternative to incorporating censored individuals, treating survival outcomes as binary ones. Forgoing the requirement of custom loss functions, CBNNs only require the use of standard components in machine learning libraries (specifically, the add layer to adjust for sampling bias and the sigmoid activation function) after case-base sampling. Due to the simplicity in its implementation and by extension user experience, CBNNs are both a user-friendly approach to data-driven, single event survival analysis and is easily extendable to any feed-forward neural network framework.

\hypertarget{data-and-code-availability-statement}{%
\subsection*{Data and code availability
statement}\label{data-and-code-availability-statement}}
\addcontentsline{toc}{section}{Data and code availability statement}

The MM and FLC datasets are available in the \textbf{survival} package in R
\citep{survpkg}. The Prostate dataset is available as part of the \textbf{asaur} package in R \citep{asaur}.
The code for this manuscript and its analyses can be found at \url{https://github.com/Jesse-Islam/cbnnManuscript}. The software package making CBNNs easier to implement can be found at \url{https://github.com/Jesse-Islam/cbnn}.

\hypertarget{acknowledgements}{%
\subsection*{Acknowledgements}\label{acknowledgements}}
\addcontentsline{toc}{section}{Acknowledgements}

We thank Dr.~James Meigs, the project leader of UM1DK078616 and R01HL151855, for his support and helpful discussions. We would also like to thank Dr.~James Hanley for his support and discussions while extending the case-base methodology.

\subsection*{Author contributions}

J.I. conceived the proof of concept, developed the theoretical formalism, developed the neural network code base, wrote the majority of the manuscript, performed the simulations and analyses. M.T. wrote the majority of the case-base sampling section \ref{case-base-sampling}. M.T. and S.B. provided key insights into the core sampling technique (case-base). R.S. provided multiple edits to the manuscript structure and key insights. All authors discussed the results and contributed to the final manuscript.

\subsection*{Financial disclosure}

The subcontracts from UM1DK078616 and R01HL151855 to R.S supported this work. The work was also supported as part of the Congressionally Directed Medical Research Programs (CDMRP) award W81XWH-17-1-0347.

\subsection*{Declaration of generative AI and AI-assisted technologies in the writing process} During the preparation of this work the author(s) used Microsoft 365 (Word) and ProWritingAid in order to improve grammar, assess typos and improve general sentence structure. After using these tools, the authors reviewed and edited the content as needed and take full responsibility for the content of the publication.

\subsection*{Conflict of interest}

The authors declare no potential conflict of interests.

 \bibliographystyle{elsarticle-harv} 
 \bibliography{manuscript}





\end{document}